\documentclass[conference]{IEEEtran}
\IEEEoverridecommandlockouts
\usepackage{cite}
\usepackage{url}
\usepackage{amsmath,amssymb,amsfonts}
\usepackage{algorithmic}
\usepackage{graphicx}
\usepackage{textcomp}
\usepackage{xcolor}
\usepackage{float}
\def\BibTeX{{\rm B\kern-.05em{\sc i\kern-.025em b}\kern-.08em
    T\kern-.1667em\lower.7ex\hbox{E}\kern-.125emX}}
\begin{document}

\title{Learning Autonomous Docking Operation of Fully Actuated
 Autonomous Surface Vessel from Expert data}

\author{\IEEEauthorblockN{ Akash Vijayakumar}
\IEEEauthorblockA{\textit{Marine Autonomous Vehicles Laboratory} \\
\textit{Department of Ocean Engineering}\\
\textit{Indian Institute of Technology Madras}\\
Chennai, India \\
akashvnkm123@gmail.com}
\and
\IEEEauthorblockN{ Atmanand M A}
\textit{Department of Ocean Engineering}\\
\IEEEauthorblockA{\textit{Indian Institute of Technology Madras  } \\
% \textit{Indian Institute of Technology Madras}\\
Chennai, India \\
atma@iitm.ac.in}
\and
\IEEEauthorblockN{ Abhilash Somayajula}
\textit{Department of Ocean Engineering}\\
\IEEEauthorblockA{\textit{Marine Autonomous Vehicles Laboratory} \\
\textit{Indian Institute of Technology Madras}\\
Chennai, India \\
abhilash@iitm.ac.in}
}
%********************************************

\maketitle

\begin{abstract}
This paper presents an approach for autonomous docking of a fully actuated autonomous surface vessel using expert demonstration data. We frame the docking problem as an imitation learning task and employ inverse reinforcement learning (IRL) to learn a reward function from expert trajectories. A two-stage neural network architecture is implemented to incorporate both environmental context from sensors and vehicle kinematics into the reward function. The learned reward is then used with a motion planner to generate docking trajectories. Experiments in simulation demonstrate the effectiveness of this approach in producing human-like docking behaviors across different environmental configurations.

\end{abstract}


\section{Introduction}

Autonomous docking of unmanned surface vessels remains a challenging problem due to the complex hydrodynamics, environmental disturbances, and constrained maneuverability. There are lots of expert data created from docking of vessels such as ferries, yachts etc in real world scenarios through Automatic Identification System (AIS) and its own logging system equipped in vessels. This data can effectively be used for teaching autonomous surface vessels (ASV) to perform autonomous operations such as docking, navigation,  collision avoidance etc. 

  Through this way the experience of human captains can be captured in ASVs to perform autonomous operations. This essentially enables such autonomous vessels to coexist with traditional vessels. Rule-based approaches typically rely on handcrafted heuristics or predefined algorithms, which may struggle to adapt to complex and dynamic environments. In contrast, Neural networks which can learn from expert demonstrations, allows the autonomous system to adapt its behavior based on real world data.

   This approach can be used in vessels such as yachts, ferries to collect data while in human operation mode and use that data for  learning operations such as docking, collision avoidance from the human expert data,  without using predefined rule based approaches. This can gradually make such vessels to learn policies for autonomous operations after it learns from human mode of operations, which can essentially lead to making those vessels autonomous. As the vessels collect more and more data, the model learns and adapts to more complex behaviours in daily operations of vessel which is out of the picture for traditional rule based approaches.

% Autonomous maritime operations have gained significant attention in recent years, driven by the potential to improve safety, efficiency, and operational costs. One critical aspect of these operations is the autonomous docking of vessels, which requires precise maneuvering in constrained environments. Traditional control methods often struggle with the complexities and uncertainties inherent in maritime environments. This paper explores the application of Inverse Reinforcement Learning (IRL) to develop a robust autonomous docking system for a fully actuated catamaran vessel.

Imitation learning approaches exists to capture the expert behaviour from data to perform that specific task. Here in this paper one such approach Inverse Reinforcement learning algorithm (IRL) has been employed to capture expert behaviour from the generated data through simulation. More specifically since the reward function which is highly non linear in nature, it can be better captured by a variant of IRL, Maximum Entropy Deep Reinforcement Learning (MEDIRL) has been used in this implementation.

This implementation  integrates both environmental context and vessel kinematics into a deep inverse reinforcement learning framework for predicting the appropriate reward function and generating a policy for docking maneuvers. By employing a two-stage neural network architecture, we effectively process and combine information about the surrounding environment (such as dock layout, occupied berths, and obstacles) with the vessel's kinematic data. This allows our system to generate docking strategies that are both safe and efficient, adapting to various scenarios much like an experienced human operator would.

\section{Related Works}
Several recent works have explored applying inverse reinforcement learning (IRL) and deep learning approaches to autonomous vehicle navigation and trajectory prediction tasks.
Wulfmeier et al. \cite{wulfmeier2017large} proposed a maximum entropy deep IRL framework for learning traversability maps from expert demonstrations in urban environments. They used a fully convolutional neural network architecture to map raw sensor inputs to reward values. Zhang et al. \cite{zhang2018integratingkinematicsenvironmentcontext} extended this work by incorporating both kinematic features and environmental context in a two-stage network to improve trajectory predictions for off-road vehicles.
For on-road autonomous driving, Fernando et al. \cite{fernando2021deep} provided an overview of deep IRL methods and demonstrated their effectiveness for long-term trajectory forecasting compared to supervised learning approaches. They highlighted the ability of deep IRL to recover underlying reward functions that explain complex driving behaviors.
In the autonomous vehicle navigation domain, Lee et al. \cite{lee2024inverse} applied IRL with dynamic occupancy grid maps to learn local path planning for autonomous vehicles in urban environments. Their CNN-based approach was able to generate collision-free trajectories by considering both static and dynamic obstacles in complex urban driving scenarios.
For modeling interactions between multiple agents, Zhao et al. \cite{zhao2019multi} proposed a tensor fusion network to capture contextual information from neighboring vehicles for trajectory prediction. Deo and Trivedi \cite{deo2018convolutional} introduced convolutional social pooling to model inter-vehicle interactions for highway driving scenarios.
Several works have also explored generative adversarial imitation learning (GAIL) for autonomous driving. Kuefler et al. \cite{kuefler2017imitating} applied GAIL with recurrent neural networks to learn human-like driving policies from demonstrations. Li et al. \cite{li2017infogail} extended GAIL with an information-theoretic regularization to learn interpretable latent factors in expert demonstrations.
The success of these approaches in related autonomous navigation domains suggests the potential for applying deep IRL and imitation learning techniques to the autonomous docking problem for marine vessels. However, the unique challenges of the maritime environment and docking maneuvers necessitate further investigation to adapt these methods effectively.

\section{Inverse Reinforcement learning }

Inverse Reinforcement learning is one of the imitation learning algorithm which uses expert data to infer a reward function and formulate a policy based on the inferred reward function. The original inverse reinforcement learning algorithm introduced has been evolved to handle from linear reward functions  to include non linear reward function through deep learning framework in past couple of years. Maximum Entropy deep inverse reinforcement learning, a variant of IRL which uses neural network to approximate the reward function, has been used  in this paper for implementation . 

\subsection{Inverse Reinforcement Learning}

In a standard Markov Decision Process (MDP), the system is characterized by a 5-tuple $(S, A, P_{sa}, R, \gamma)$, where $S$ denotes the set of states, $A$ represents the set of possible actions, $P_{sa}$ defines the state transition probabilities, $R$ is the immediate expected reward upon transitioning between states, and $\gamma$ is the discount factor applied to future rewards. The goal of Inverse Reinforcement Learning (IRL) is to infer the reward function $R$ from a set of $N$ expert demonstrations $D = (\zeta_1, \zeta_2, \dots, \zeta_N)$, where each demonstration $\zeta_i$ consists of a sequence of states $\{s_1, \dots, s_T\}$, with $T$ representing the length of the trajectory. Ng and Russell \cite{ng2000algorithms} proposed a framework where the reward $R$ is expressed as a linear function of state features, parameterized by $\theta$: $R_\theta(s) = \theta^T f(s)$, where $\theta \in \mathbb{R}^n$ is the parameter vector and $f(s): S \rightarrow \mathbb{R}^n$ maps states to feature vectors. Given the discount factor $\gamma$ and a policy $\pi$, the reward function $R$ is defined as the expected cumulative discounted reward:

\begin{equation}
    \mathbb{E}\left[\sum_{t=0}^{\infty} \gamma^t R_\theta(s_t) \mid \pi\right] = \mathbb{E}\left[\sum_{t=0}^{\infty} \gamma^t \theta^T f(s_t) \mid \pi\right] = \theta^T \bar{f}(\pi)
\end{equation}

where $\bar{f}(\pi)$ represents the expected cumulative discounted feature values, or feature expectations . Abbeel and Ng \cite{abbeel2004apprenticeship} demonstrated that if the feature expectations of the expert and the learner align, the learner's policy will perform as well as the expert's policy.

\subsection{Maximum Entropy Inverse Reinforcement Learning}

When expert demonstrations are imperfect or noisy, representing the behavior with a single reward function becomes challenging. Ziebart et al. \cite{ziebart2008maximum} introduced the Maximum Entropy IRL (MaxEnt IRL) framework to address this issue. By maximizing the entropy of path distributions while ensuring the feature expectation matching constraints \cite{abbeel2004apprenticeship}, MaxEnt IRL maximizes the likelihood of the observed data $D$ under the assumed maximum entropy distribution \cite{ziebart2008maximum}:

\begin{equation}
    \theta^* = \arg\max_{\theta} L(\theta) = \arg\max_{\theta} \sum_{\zeta \in D} \log P(\zeta \mid \theta, P_{sa})
\end{equation}

Here, $P(\zeta \mid \theta, P_{sa})$ follows the maximum entropy (Boltzmann) distribution \cite{ziebart2008maximum}. This convex optimization problem is solved using gradient-based methods, where the gradient is given by:

\begin{equation}
    \frac{\partial L(\theta)}{\partial \theta} = \sum_{s \in \zeta \in D} \mu_s f(s) - \sum_{s_i} \mu_{s_i} f(s_i)
\end{equation}

In this context, $\mu_s$ represents the State Visitation Frequency (SVF), which is the discounted sum of probabilities of visiting state $s$: 

\begin{equation}
    \mu_s = \sum_{t=0}^{\infty} \gamma^t P(s_t = s \mid \pi, \theta, P_{sa})
\end{equation}

With a specified feature function $f$, this update rule iteratively adjusts $\theta$ to align the optimal policy's SVF with that of the expert demonstrations $D$.

\subsection{Maximum Entropy Deep Inverse Reinforcement Learning}

Traditional approaches to estimating the reward function often rely on a linear combination of manually selected features. To overcome the limitations of this linear approach, Wulfmeier et al. \cite{wulfmeier2015maximum} proposed using neural networks to generalize the reward function to a nonlinear form, $R_\theta(s) = R(f(s), \theta)$. By training a neural network (NN) with raw sensory data as input, both the weights and features are automatically learned, eliminating the need for manually designed state features. In Maximum Entropy Deep Inverse Reinforcement Learning (MEDIRL), the network is trained to maximize the joint probability of the demonstration data $D$ and the model parameters $\theta$ under the learned reward function $R_\theta(s)$:

% \begin{equation}
%     L(\theta) = \log P(D, \theta \mid R_\theta(s)) = \log P(D \mid R_\theta(s)) + \log P(\theta) = L_D + L_\theta
% \end{equation}

\begin{equation}
\begin{split}
    L(\theta) &= \log P(D, \theta \mid R_\theta(s)) \\
    &= \log P(D \mid R_\theta(s)) + \log P(\theta) \\
    &= L_D + L_\theta
\end{split}
\end{equation}

Here, $L_\theta$ can be optimized using weight regularization techniques common in NN training, allowing MEDIRL to focus on maximizing the first term $L_D$:

\begin{equation}
    \frac{\partial L_D}{\partial \theta} = \frac{\partial L_D}{\partial R_\theta} \frac{\partial R_\theta}{\partial \theta} = (\mu_D - \mathbb{E}[\mu]) \frac{\partial R_\theta(s)}{\partial \theta}
\end{equation}

In this context, $\mathbb{E}[\mu]$ represents the expected SVF derived from the predicted reward. The MEDIRL update (6) allows for the straightforward computation of the gradient of the reward with respect to the weight parameters through backpropagation \cite{ziebart2010modeling}.

In this paper the maximum entropy deep IRL framework used resemble the implementation of Zhang et al\cite{zhang2018integratingkinematicsenvironmentcontext}. The network has  a convolution block to extract features and another convolution block to extract reward map from integrated feature maps and kinematics.

% \bibliographystyle{IEEEtran}
% \bibliography{references}

\section{Simulation}

For training the MEDIRL algorithm, expert data is required so that the docking vessel can imitate the expert behaviour. In this paper a docking simulation setup has been implemented and data is generated through a sampling based RRT* planning algorithm .  

In the docking task, the environment is set up with eight potential docking bays, each measuring 3m by 3m, where the vessel can dock. These docking bays are divided by walls, with four docks on one side and the remaining four directly opposite, separated by an 8m-wide waterway. Rectangular piers positioned between the docks serve as static obstacles. At the start of each simulation, four out of the eight docks are randomly selected and occupied by other vessels, preventing the docking vessel from using these occupied bays. The docking vessel operates with three degrees of freedom: surge, sway, and yaw. It is spawned at a random location near the docking area each time the simulation runs. The goal is to dock at the nearest unoccupied bay. A path is planned using the RRT* algorithm with 10,000 iterations, treating the vessel as a point object with a collision checker matching its dimensions. The planned trajectory is then executed by a PD controller. During the execution, trajectory data, including x and y coordinates along with environmental information, is recorded to serve as expert data for training the inverse reinforcement learning algorithm.

\begin{figure}[h]
    \centering
    \includegraphics[width=0.5\textwidth]{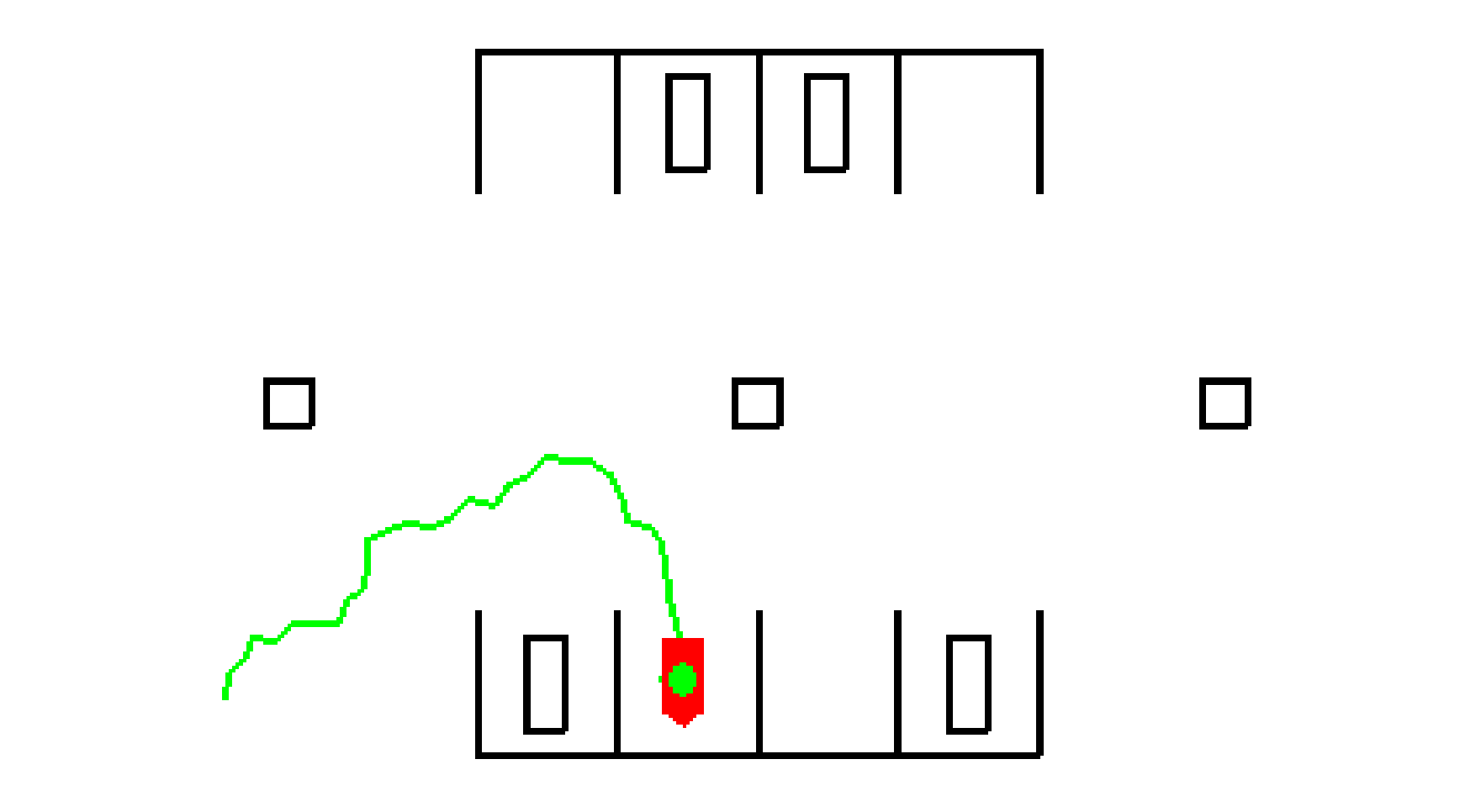}
    \caption{Simulation Environment}
    \label{fig:example}
\end{figure}

\section{Feature Extraction and Kinematic Regression }

\begin{figure*}[t]
    \centering
    \includegraphics[width=0.7\textwidth]{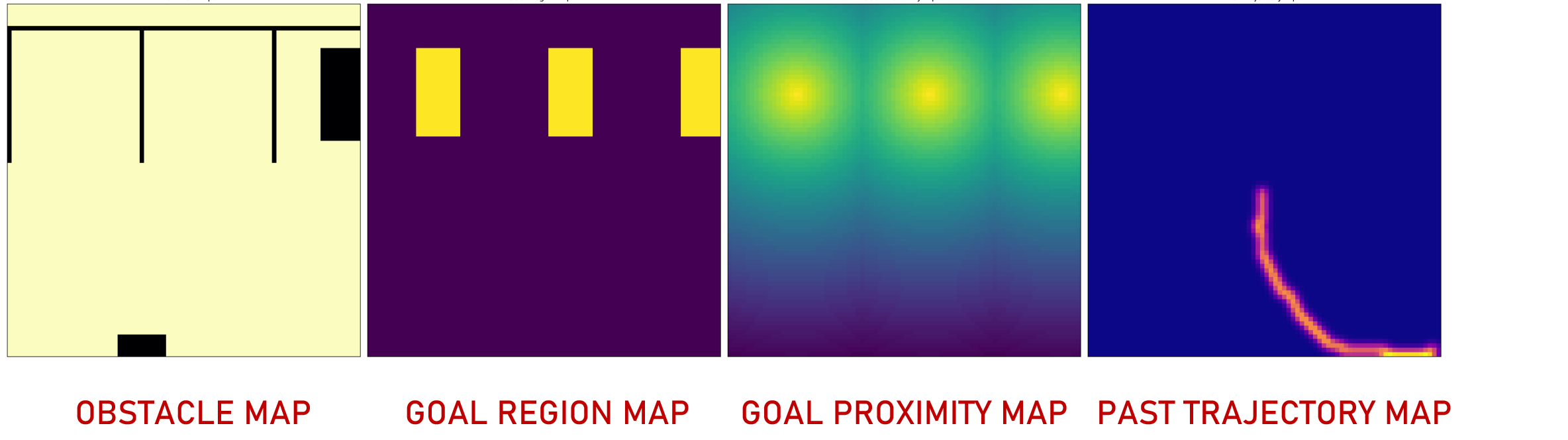}
    \caption{Input Feature Maps}
    \label{fig:featinmaps}
    
\end{figure*}

\begin{figure*}[t]
    \centering
    \includegraphics[width=0.7\textwidth]{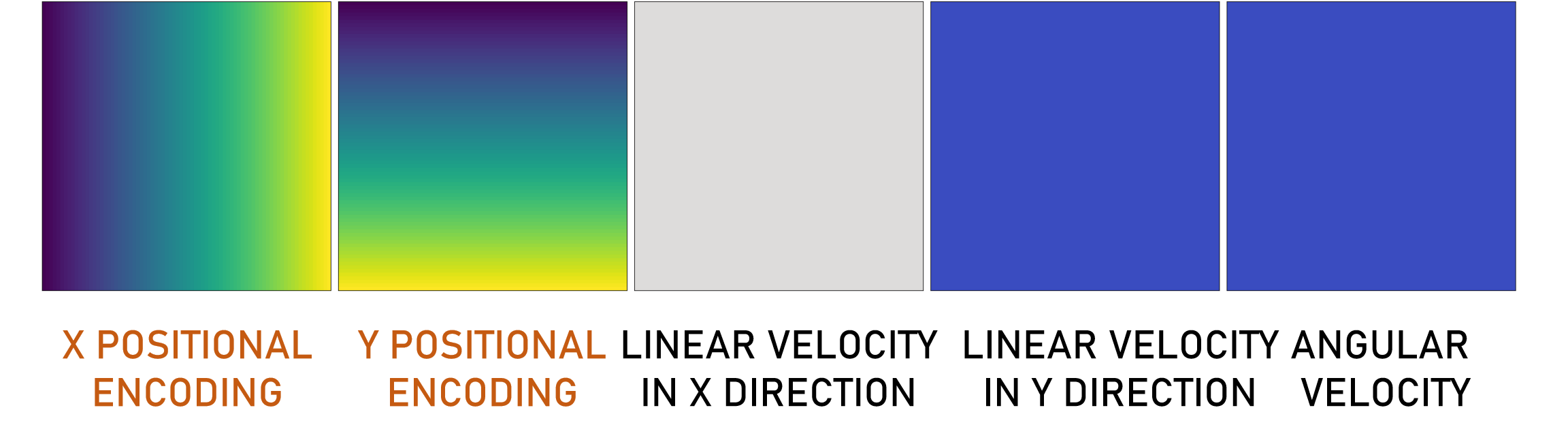}
    \caption{Input Kinematics Feature Maps}
    \label{fig:kinfeatinmaps}
    
\end{figure*}

\begin{figure}[H]
    \centering
    \includegraphics[width=0.3\textwidth]{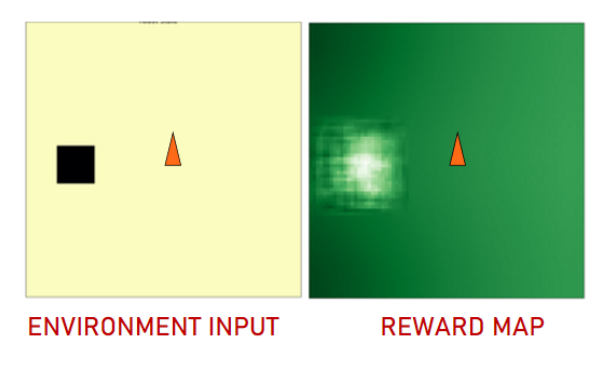}
    \caption{Input Environmental Map and Reward Map from Network}
    \label{fig:Rewardmap}
\end{figure}

\begin{figure*}[h]
    \centering
    \includegraphics[width=1\textwidth]{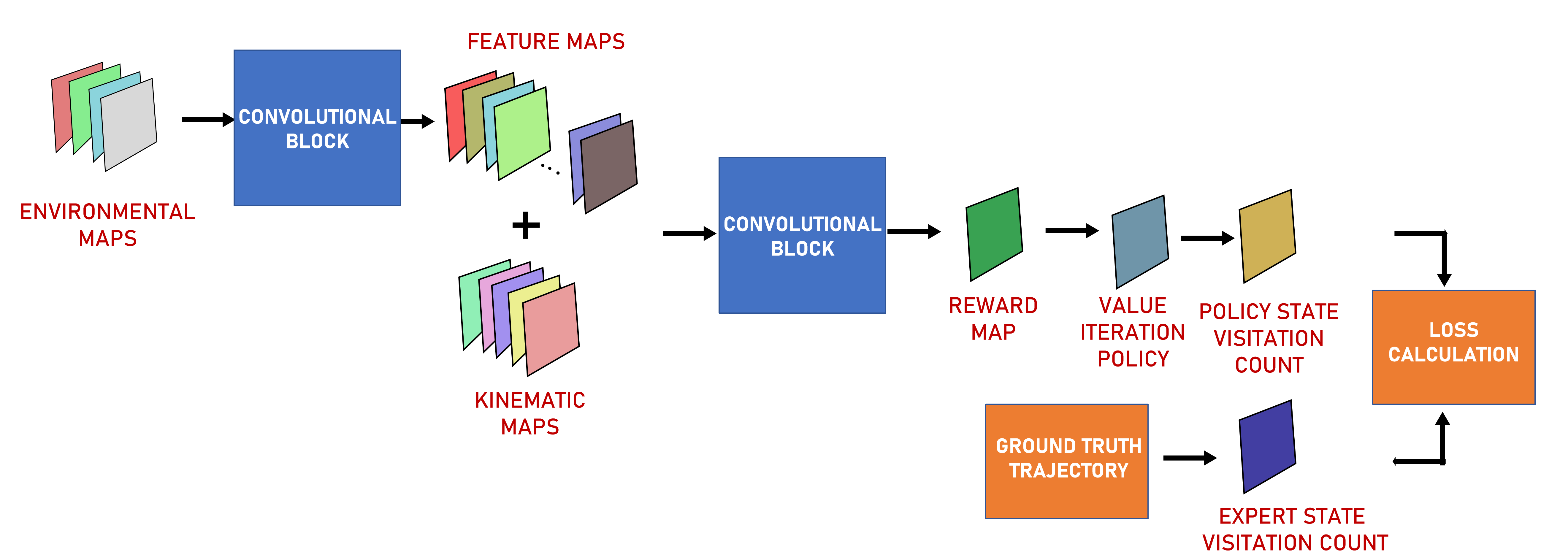}
    \caption{Network Architecture}
    \label{fig:medirlnet}
\end{figure*}

The network architecture employed in this experiment closely resembles the one described in the implementation of Zhang et. al \cite{zhang2018integratingkinematicsenvironmentcontext}. The input consists of feature maps that include the environment information map, goal proximity map, goal region map, and past trajectory map \ref{fig:featinmaps} as well as kinematic features. Each feature is generated within a 4m*4m vessel centered grid at that time step, which necessarily capture sufficient information on its current state and surroundings. A convolutional block is used to extract feature representations from these input maps. These extracted features are then combined with kinematic data, such as velocity in the X and Y directions, angular velocity, and positional encoding maps for both X and Y coordinates. The X and Y positional encoding maps essentially represents the X, Y coordinates of the grid cells of vessel centered grid. The kinematics input which includes linear velocity in X and Y , angular velocity of the vessel are uniform across the grid cells.\ref{fig:kinfeatinmaps}  The final output of the network is treated as a reward map as in figure \ref{fig:Rewardmap}, which is used in the back propagation process during inverse reinforcement learning training as shown in network architecture in figure \ref{fig:medirlnet}.

\section{Results and Discussion}

\begin{figure}[H]
    \centering
    \includegraphics[width=0.3\textwidth]{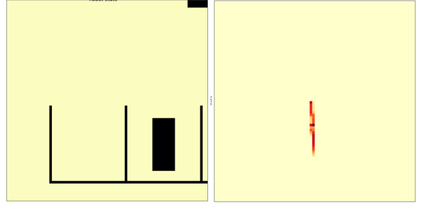}
    \caption{Go Forward Affinity}
    \label{fig:goforward}
\end{figure}

\begin{figure}[H]
    \centering
    \includegraphics[width=0.3\textwidth]{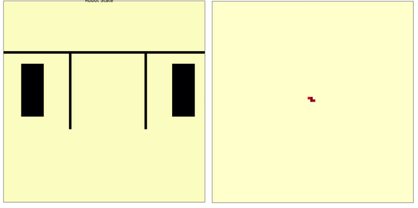}
    \caption{Inside Dock}
    \label{fig:insidedock}
\end{figure}

\begin{figure}[H]
    \centering
    \includegraphics[width=0.3\textwidth]{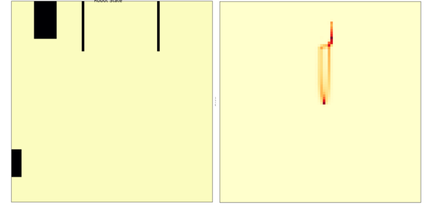}
    \caption{Two branches}
    \label{fig:twobranches}
\end{figure}

The network has been trained on 500 trajectories generated by the simulation and tested on 50 different trajectories . The test results shows that the model is able to capture the environmental context as well as the kinematics of the vessel to generate possible path from its current position. The results shown such that  the environmental information map from the vessel centered frame and the state visitation map generated by the trained policy are shown side by side. The environmental information map shows where the obstacle around are located in a vessel centred frame. The state visitation map conveys the probable path of the vessel considering the current state information.  In the figure \ref{fig:goforward} it has been seen that the vessel stays in front of the dock has strong affinity to go towards the empty dock and it also generates a path to the leftwards with lesser affinity since its occupied by another vessel. In figure \ref{fig:twobranches} the vessel is in front and almost middle of an occupied and empty docks, the policy generates two probable path towards the goal position which terminally direct towards the empty dock . In figure \ref{fig:insidedock} the vessel is perfectly inside the dock and the policy generates a dot which signifies the docked position of the vessel which further requires no path to destination.   

% \begin{figure}[hb]
%     \centering
%     \includegraphics[width=0.4\textwidth]{chosenresultimgs/equiprobablecrop.png}
%     \caption{Equi Probable}
%     \label{fig:equiprob}
% \end{figure}

\section{Conclusion}

The proposed approach of leveraging inverse reinforcement learning to enable autonomous docking of unmanned surface vessels from expert demonstrations holds significant
promise. However, there are numerous avenues for further development and exploration of this research. Currently the environment handles static obstacles it can be extended to dynamic obstacles using recurrent neural networks which is an avenue of further development. Extending the framework to handle multi-agent coordination and docking
scenarios could unlock new possibilities for coordinated maritime operations and cooperative robotic missions. Furthermore, leveraging transfer learning and knowledge sharing could accelerate the adaptation process to new environments or vessel configurations. Finally, the principles and techniques developed in this research could potentially be extended to other maritime or robotic tasks that require learning complex behaviors from expert demonstrations, such as
autonomous navigation, obstacle avoidance, or specialized missions like search and rescue, environmental monitoring.By pursuing these avenues, this research could pave the way for advanced autonomous maritime operations, improved safety and efficiency in in the maritime domain.

% \bibliographystyle{asmeconf}

% \bibliography{oceans}


% \bibitem{b1} G. Eason, B. Noble, and I. N. Sneddon, ``On certain integrals of Lipschitz-Hankel type involving products of Bessel functions,'' Phil. Trans. Roy. Soc. London, vol. A247, pp. 529--551, April 1955.
% \bibitem{b2} J. Clerk Maxwell, A Treatise on Electricity and Magnetism, 3rd ed., vol. 2. Oxford: Clarendon, 1892, pp.68--73.

% \end{thebibliography}

\end{document}